\documentclass[conference]{ieeeconf}

\usepackage{graphicx}
\usepackage{bm}
\usepackage{amsmath}
\usepackage{amssymb}
\usepackage{amsbsy}
\usepackage[noadjust]{cite}
\usepackage{xcolor}
\usepackage[ruled,vlined,linesnumbered]{algorithm2e}
\usepackage[absolute]{textpos}

\IEEEoverridecommandlockouts
\renewcommand{\vec}[1]{\boldsymbol{\mathrm{#1}}}

\TPMargin{5pt}

\newcommand{\copyrightstatement}{
    \begin{textblock}{0.84}(0.08,0.93)    
         \noindent
         \scriptsize
         \copyright 2021 IEEE.  Personal use of this material is permitted.  Permission from IEEE must be obtained for all other uses, in any current or future media, including reprinting/republishing this material for advertising or promotional purposes, creating new collective works, for resale or redistribution to servers or lists, or reuse of any copyrighted component of this work in other works.
    \end{textblock}
}
\newcommand{\versionstatement}{
    \begin{textblock}{0.84}(0.08,0.04)    
         \noindent
         \raggedleft
         Accpeted version for publication at IEEE EMBC 2021
    \end{textblock}
}

\title{\LARGE \bf
    On the interpretation of linear Riemannian tangent space model parameters in M/EEG
}

\author{Reinmar J. Kobler$^{1,2}$, Jun-Ichiro Hirayama$^{1}$, Lea Hehenberger$^{2,3}$, Catarina Lopes-Dias$^{2,3}$, \\Gernot R. M{\"u}ller-Putz$^{2,3}$ and Motoaki Kawanabe$^1$
    \thanks{$^1$RIKEN AIP, Kyoto, Japan.%
    ~$^2$Graz BCI Racing Team, $^3$Insitute of Neural Engineering, Graz University of Technology, Graz, Austria.~%
    E-mail: {\tt\footnotesize \{reinmar.kobler, motoaki.kawanabe\}@riken.jp}}%
}

\begin{document}
\bstctlcite{IEEEexample:BSTcontrol}
    \maketitle
    \thispagestyle{empty}
    \pagestyle{empty}

    \pagestyle{plain}
    \versionstatement
    \copyrightstatement
    
    \begin{abstract}
        Riemannian tangent space methods offer state-of-the-art performance in magnetoencephalography (MEG) and electroencephalography (EEG) based applications such as brain-computer interfaces and biomarker development.
        One limitation, particularly relevant for biomarker development, is limited model interpretability compared to established component-based methods.
        Here, we propose a method to transform the parameters of linear tangent space models into interpretable patterns.
        Using typical assumptions, we show that this approach identifies the true patterns of latent sources, encoding a target signal.
        In simulations and two real MEG and EEG datasets, we demonstrate the validity of the proposed approach and investigate its behavior when the model assumptions are violated.
        Our results confirm that Riemannian tangent space methods are robust to differences in the source patterns across observations.
        We found that this robustness property also transfers to the associated patterns. 
    \end{abstract}

    \section{Introduction}
    \noindent
    Magnetoencephalography (MEG) and electroencephalography (EEG) capture a linear mixture of brain and noise signals \cite{nunez_electric_2006}.
    In a supervised setting, where the goal is to infer a target signal from the power of latent oscillatory sources, 
    component-based methods like common spatial patterns (CSP) \cite{ramoser_optimal_2000,Blankertz2008} for discrete targets or source power co-modulation (SPOC) \cite{Dahne2014} for continuous targets are widely used \cite{lotte_review_2018}.
    Recently, they have been outperformed by Riemannian tangent space methods in several datasets \cite{congedo_riemannian_2017,gemein_machine-learning-based_2020,sabbagh_predictive_2020,xu2020tangent}.
    
    Key factors for the success of Riemannian tangent space methods are that the features, namely covariance matrices, lie on a Riemannian manifold and the commonly used geometric metric is invariant to affine transformations \cite{yger_riemannian_2017,congedo_riemannian_2017}.
    The tangent space is a vector space with a Euclidean metric that locally approximates the Riemannian manifold around a reference matrix, typically the geometric mean of a dataset.
    Consequently, standard linear machine learning techniques for Euclidean vector spaces can be used in the tangent space.

    One limitation of the Riemannian approaches is their lack of interpretability in terms of contributing brain sources \cite{sabbagh_manifold-regression_2019,xu2020tangent}.
    For component-based methods, there are established techniques to interpret the model parameters in terms of spatial patterns \cite{Haufe2014}.
    In this work, we show that the parameters of linear regression and classification methods in the Riemannian tangent space can be transformed to interpretable spatial patterns in the M/EEG channel space in a similar fashion as for component-based methods.
    Thereby, the tradeoff between performance and interpretability can be overcome.

    In the next section, we introduce the underlying generative model for a regression problem, briefly outline a recently proposed tangent space regression algorithm \cite{sabbagh_predictive_2020}, followed by the proposed method to convert the parameters to interpretable patterns.
    The section ends with a description of conducted simulations, analyzed datasets, and baseline methods.

    \section{Materials and Methods}
    \noindent

    \subsection{Generative model}
    \noindent
    The M/EEG signals $\vec{x}_i(t)$ are typically modelled as a linear mixture of sources plus additive noise \cite{nunez_electric_2006}
    \begin{equation}
        \vec{x}_i(t) = \vec{A}_s \vec{s}_i(t) + \vec{n}_i(t)
        \label{eq:genmdl}
    \end{equation}
    where $\vec{s}_i(t) \in \mathbb{R}^Q$ denotes the source signal time activity of observation $i$ (epoch, session, subject, etc.) and $\vec{n}_i(t) \in \mathbb{R}^P$ the additive noise.
    The matrix $\vec{A}_s\in \mathbb{R}^{P \times Q}$ contains the $Q$ source patterns \cite{Haufe2014}.
    
    As in \cite{sabbagh_predictive_2020}, we assume that the brain signals arise from activity of uncorrelated sources.
    The noise $\vec{n}_i(t) = \vec{A}_n \vec{\nu}_i(t)$ is stationary, uncorrelated with the sources, and spans a subspace ($ \vec{A}_n \in \mathbb{R}^{P \times P-Q}$) that is shared across observations.
    The generative model can then be written as:
    \begin{equation}
        \vec{x}_i(t) = \vec{A}_s \vec{s}_i(t) + \vec{A}_n \vec{\nu}_i(t) = \vec{A} \vec{\eta}_i(t)
    \end{equation}
    $\vec{A}\in \mathbb{R}^{P \times P}$ includes the source and noise patterns and is assumed to be invertible.
    The vector $\vec{\eta}_i(t) \in \mathbb{R}^P$ is the concatenation of the latent source and noise signals.
    A scalar target signal $y_i$ at observation $i$ is then modeled as a function of the latent sources' powers:
    \begin{equation}
        y_i = \vec{b}^T f(\vec{p}_i) + b_0 + \varepsilon_i
        \label{eq:regmdl}
    \end{equation}
    where $\vec{p}_i = \mathbb{E}\{\vec{s}_i^2(t)\}_t \in \mathbb{R}^Q$ contains the sources' powers at observation $i$, $f(\cdot)$ is a known function, $\vec{b} \in \mathbb{R}^Q$ a weight vector, $b_0$ a bias term, and $\varepsilon_i \sim \mathcal{N}(0, \sigma^2)$ noise.
    Here, we consider $f(\cdot) = \log(\cdot)$ as log-linear relationships are often encountered in oscillatory brain activity \cite{buzsaki_log-dynamic_2014}.

    Given a set of paired observations $ \{(\vec{x}_i(t), y_i)_{i=1,..,N}\} $ our goal is to predict the target signal for unseen data, and identify the patterns $\vec{A}_s$ of the encoding sources.

    As in previous works \cite{sabbagh_predictive_2020,Dahne2014}, we will use the between-sensor covariance matrices as features.
    Assuming zero-mean signals, the covariance matrices can be computed as:
    \begin{equation}
        \vec{C}_i = \frac{1}{T} \vec{X}_i \vec{X}_i^T \in \mathbb{R}^{P \times P}
    \end{equation}
    where the columns of the matrix $\vec{X}_i \in \mathbb{R}^{P \times T}$ contain $T$ temporal samples.
    The covariance matrices $\vec{C}_i$ are in the manifold of positive definite matrices $ \mathcal{S}_P^{++}$.
    If we assume that the source signals are zero-mean and uncorrelated, their covariance matrix is diagonal $\mathbb{E}\{\vec{s}_i(t) \vec{s}_i^T(t)\}_t = \mathrm{diag}\left(\vec{p}_i\right)$.
    If they are also uncorrelated with the noise, i.e. $\mathbb{E}\{\vec{s}_i(t) \vec{\nu}_i^T(t)\}_t = 0$, 
    and w.l.o.g. the noise sources are uncorrelated, 
    the sensor covariance matrices can be expressed as:
    \begin{equation}
        \vec{C}_i = \vec{A} \vec{E}_i \vec{A}^{T}
        \label{eq:relationbetween_ci_and_ei}
    \end{equation}
    where $\vec{E}_i = \mathbb{E}\{\vec{\eta}_i(t) \vec{\eta}_i^T(t)\}_t$ is a diagonal matrix, whose diagonal elements are $\vec{p}_i$.

    \subsection{Riemannian tangent space regression model}
    \noindent
    Equipping the manifold $\mathcal{S}_P^{++}$ with the geometric metric gives a Riemannian manifold structure to $\mathcal{S}_P^{++}$.
    For the generative model, defined in (\ref{eq:genmdl}), the Riemannian tangent space embedding of the covariance matrices $\vec{C}_i$ gives a consistent estimator for $y_i$, if the function $f(\cdot)$ is the logarithm \cite{sabbagh_predictive_2020}.
    The embedding $\vec{v}_i$ is computed as:
    \begin{equation}
        \vec{v}_i = \mathrm{proj}_{\bar{\vec{C}}} (\vec{C}_i) = \mathrm{upper}\left( \mathrm{log}\left( \bar{\vec{C}}^{-1/2} \vec{C}_i \bar{\vec{C}}^{-1/2} \right) \right)
        \label{eq:C_i to v_i}
    \end{equation}
    where $\bar{\vec{C}}$ is the geometric mean \cite{congedo_riemannian_2017} of the covariance matrices $\{\vec{C}_1,..,\vec{C}_N\}$, 
    the logarithm for $\vec{C} \in\mathcal{S}_P^{++}$ is $\mathrm{log}(\vec{C}) = \vec{U} \mathrm{diag}(\mathrm{log}(\lambda_1), .., \mathrm{log}(\lambda_P)) \vec{U}^T$ with $\vec{U}^T \vec{U} = \vec{I}$,
    and the invertible mapping $\mathrm{upper}(\vec{C}) \in \mathbb{R}^{P(P+1)/2} $ extracts the upper triangular elements of a symmetric matrix, with the off-diagonal elements weighted by the factor $\sqrt{2}$.
    The weighting ensures that $||\mathrm{upper}(\vec{C})||_2 = ||\vec{C}||_F$.
    The Euclidean distance $||\vec{v}_k - \vec{v}_l||_2$ in the tangent space approximates the geometric distance between $\vec{C}_k$ and $\vec{C}_l$ in the vicinity of $\bar{\vec{C}}$ \cite{congedo_riemannian_2017}.
    
    Since the relation between $\vec{v}_i$ and $y_i$ is linear \cite{sabbagh_predictive_2020}, any linear method can be used to fit $\vec{b}_c$ so that a cost function between $\hat{y}_i = \vec{b}_c^T \vec{v}_i  + b_0$ and $y_i$ is minimized.
    For M/EEG datasets the observation ($N$) to feature ($P(P+1)/2$) ratio is typically small, requiring regularization.

    \subsection{M/EEG channel space model patterns}
    \label{sec:pattern computation}
    \noindent
    Given a fitted model with parameters $(\vec{b}_c,\,\bar{\vec{C}})$, the patterns $\vec{A}_s$ of the encoding sources can be identified in a three step procedure (algorithm\,\ref{algo:patterns}).
    First, the tangent space pattern $\vec{d}_c$ associated to $\vec{b}_c$ is computed according to \cite{Haufe2014},
    using $\vec{C}_v = \mathbb{E}\{\vec{v}_i \vec{v}_i^T\}_i$.
    Next, $\vec{d}_c$ is projected back to the covariance matrix space to obtain $\vec{C}_d \in \mathcal{S}_P^{++}$.
    Finally, the general eigenvalue problem for $\vec{C}_d$ and $\bar{\vec{C}}$ is solved.
    
    Under the generative model, the resulting eigenvectors correspond to the patterns $\vec{A}$ and the eigenvalues $\left(\lambda_j\right)_{j=1,..,P}$ are a function of the unknown, weight vector $\vec{b}$ (see proof in the appendix).
    Specifically, the eigenvalues are:
    \begin{equation}
        \lambda_j (\vec{b}) =  
        \begin{cases}
            \mathrm{exp}\left(b_j/||\vec{b}||^2\right) & j \le Q\\
            1 & otherwise
        \end{cases}
    \end{equation}
    where the first $Q$ eigenvalues correspond to the encoding sources $\vec{s}(t)$.
    In practice, the sources with the strongest coupling, i.e., largest $|b_j|$, are found via sorting the eigenvalues according to the criterion $\mathrm{max}\left(\lambda_j, 1/\lambda_j\right)$.
    The number of sources with significant coupling ($Q$) can be identified via a shuffling procedure.
    \vspace{-6pt}
    \begin{algorithm}
        \SetAlgoLined
        \SetKwInOut{Input}{Input}\SetKwInOut{Output}{Output}
        \Input{tangent space weight vector $\vec{b}_c$, \\tangent space projection matrix $\bar{\vec{C}}$, \\tangent space feature covariance matrix $\vec{C}_v$}
        \Output{$\vec{A} \in \mathbb{R}^{P\times P} $, $\vec{\lambda} \in \mathbb{R}^P$}
        \BlankLine
        $\vec{d}_c = \vec{C}_v \vec{b}_c \sigma_{\hat{y}}^{-2}$ with $\sigma_{\hat{y}}^2 = \vec{b}_c^T \vec{C}_v  \vec{b}_c $\\
        $\vec{C}_d = \mathrm{proj}_{\bar{\vec{C}}}^{-1} (\vec{d}_c) = \bar{\vec{C}}^{1/2} \mathrm{exp}\left( \mathrm{upper}^{-1}\left(  \vec{d}_c  \right) \right) \bar{\vec{C}}^{1/2}$\\
        $\vec{A}, \vec{\lambda} = \mathrm{eigh}(\vec{C}_d, \bar{\vec{C}}) $

        \caption{Channel space model patterns}
        \label{algo:patterns}
    \end{algorithm}
    \vspace{-12pt}
    \subsection{Model fitting and evaluation}
    \noindent
    In real M/EEG datasets, the covariance matrices can be rank-deficient ($ \vec{C}_i \notin  \mathcal{S}_P^{++}$). 
    In this case, the geometric metric is not defined \cite{congedo_riemannian_2017}.
    As a remedy, we reduced the dimensions from $P$ to $K$ via projecting the covariance matrices to the subspace spanned by the first $K$ principal components of the covariance matrices' arithmetic mean \cite{sabbagh_manifold-regression_2019}.
    In this subspace, we computed the geometric mean and the tangent space features, according to (\ref{eq:C_i to v_i}).
    Next, the features were z-scored.
    Depending on the dataset, the linear weight vector $\vec{b}_c$ was estimated either via ridge regression or penalized logistic regression.
    The associated cost functions were the mean absolute error (MAE) between $y_i$ and $\hat{y}_i$ or the balanced classification accuracy.
    The train/test-splitting scheme depended on the dataset.
    All parameters (spatial filters, geometric mean, z-scoring, linear weights) were fitted to training data.
    The optimal regularization parameter of the regression/classification model was determined using an inner generalized cross-validation (CV) scheme.
    We considered 25 candidate values which were log-spaced in the range $[10^{-5}, 10^{3}]$.
    In the next sections, we refer to this sequence of operations as the RIEMANN pipeline.

    To put the results into context, we compared the performance to SPOC/CSP and a na{\"i}ve method, denoted DIAG here.
    For SPOC, we used the SPOC lambda algorithm \cite{Dahne2014} to estimate $k$ components.
    After spatial filtering with SPOC or CSP, the logarithm of the covariance matrices' diagonal elements were the features.
    The subsequent steps (z-scoring, regression/classification) were identical.
    The DIAG pipeline was identical to the SPOC/CSP pipeline except for the initial spatial filtering step.
    Consequently, the method computed log-band power features in channel space.
    We computed patterns for the RIEMANN and SPOC/CSP pipelines according to section \ref{sec:pattern computation} and \cite{Haufe2014}.

    \subsection{Experiments}
    \label{sec:experiments}
    \noindent
    We conducted three experiments to demonstrate the validity of the approach and analyzed the behavior when the model assumptions are violated.

    \subsubsection{Simulations}
    \noindent
    In two regression problem simulations we investigated the algorithms' properties in identifying a single encoding source.
    First, we varied the power $\sigma^2$ of the additive Gaussian noise $\varepsilon_i$ in (\ref{eq:regmdl}).
    Second, we introduced a model violation by making the patterns dependent on the observation index $\vec{A}_i = \vec{A} + \vec{N}_i$ with $(\vec{N}_i)_{jk} \sim \mathcal{N}(0, \alpha^2)$.
    The patterns were computed as $\vec{A} = \exp(\vec{B})$ with $(\vec{B})_{jk} \sim \mathcal{N}(0, 1)$.
    The other parameters were identical to \cite{sabbagh_predictive_2020}.
    Because the covariance matrices had full rank, we omitted the PCA step for the RIEMANN pipeline.

    Ten-fold CV was used to fit and evaluate the models.
    In addition to the MAE cost function between $y_i$ and $\hat{y}_i$, we computed distances between the true $\vec{a}_s$ and its estimate $\hat{\vec{a}}_s$.
    The pattern distance was defined as $1 - |\hat{\vec{a}}_s^T \vec{a}_s| / (||\hat{\vec{a}}_s|| \cdot ||\vec{a}_s||) $.
    A distance of 0 means that the topographies are identical up a scalar factor $\gamma \in \mathbb{R}$. 
    \begin{figure}[t]
        \centering
        \includegraphics[width=\columnwidth]{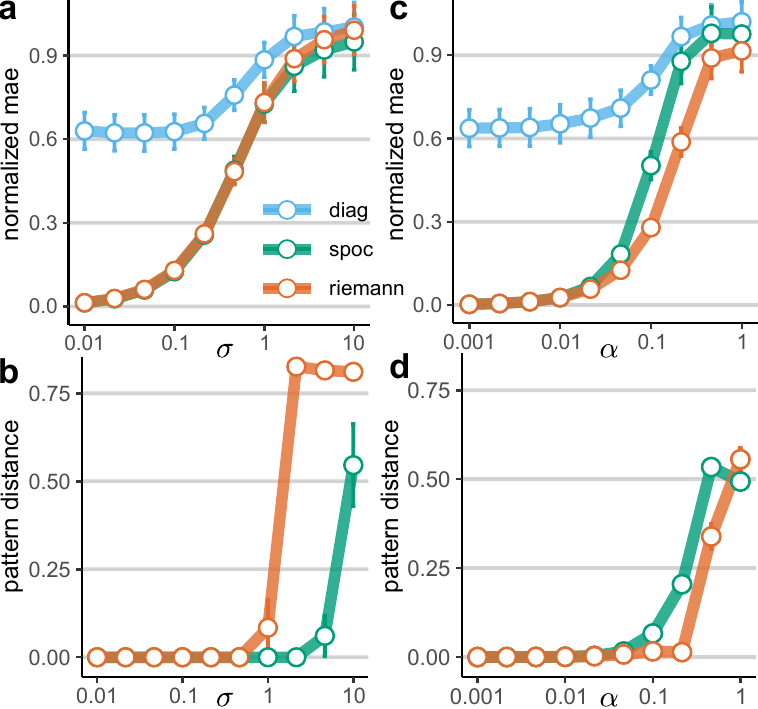}
        \caption{\textbf{Simulation results.}
                 Effect of the target noise power ($\sigma^2$) on the mean absolute error (MAE) (\textbf{a}) and the pattern distance (\textbf{b}).
                 The methods are color-coded. 
                 Error bars show the standard error of the mean.
                 \textbf{a}, The MAE was normalized by the MAE of a dummy method that used the expected value of the target signal $\mathbb{E}\{y_i\}_i$ as predictor.
                 \textbf{c} and \textbf{d}, as in \textbf{a} and \textbf{b} for different pattern noise power ($\alpha^2$) levels.} 
        \label{fig:simulation}
    \end{figure}
    \subsubsection{Cortico-muscular coherence dataset}
    \noindent
    We analyzed an MEG dataset that studied cortico-muscular coherence (CMC) \cite{schoffelen_selective_2011}.
    The publicly available dataset contains recordings of a single participant during one session.
    In a trial-based task, the participant contracted her left hand and exerted a constant force against a lever.
    The trials were interleaved with short breaks.
    As in \cite{sabbagh_predictive_2020}, we analyzed the dataset in a continuous setting with the goal to decode the EMG envelope from the MEG beta band activity of 151 gradiometers.
    The considered data and preprocessing steps were similar to \cite{sabbagh_predictive_2020}.
    In a nutshell, we set the single bipolar EMG channel power as target signal $y_i$, and
    extracted oracle approximating shrinkage (OAS) regularized \cite{chen_shrinkage_2010} covariance matrices $\vec{C}_i$ for beta band ($[15, 30]\, Hz$) activity in overlapping windows ($T = 1.5\,s$, overlap = $1.25\,s$).
    We applied 10-fold CV to evaluate the goodness of fit and used the coefficient of determination $R^2$ as metric.

    \subsubsection{Multi-session BCI dataset}
    \noindent
    This dataset was recorded during a longitudinal (26 sessions, 15 months) BCI study with a tetraplegic user \cite{hehenberger_long-term_2021}.
    The analyzed data contains EEG signals (32 channels), recorded during a trial-based paradigm.
    In each trial, the user performed 1 of 4 distinct mental tasks and received discrete feedback, provided by an adaptive BCI.
    Here, we analyzed the two tasks with the strongest patterns (feet motor imagery and mental subtraction).
    The data preprocessing and trial rejection methods were identical to \cite{hehenberger_long-term_2021}.
    The preprocessed and cleaned data comprised activity in 4 frequency bands during a 2-s epoch per trial (1438 trials).
    For each epoch and band, one OAS regularized covariance matrix $\vec{C}_i$ was computed.
    The tangent space projection was computed independently for each frequency band.
    Thereafter, the individual feature vectors were concatenated and used to predict the target class.
    The models were evaluated using a leave-one-session-out CV scheme.

    \subsection{Software}
    \noindent
    The software and analysis scripts are publicly available {\small{\texttt{https://github.com/rkobler/interpret\_ lin\_rts\_mdls}}}, and are based on the code of \cite{sabbagh_predictive_2020} and the python packages \texttt{Scikit-Learn} \cite{pedregosa2011scikit}, \texttt{MNE} \cite{gramfort_mne_2014} and \texttt{PyRiemann} \cite{barachant_multiclass_2012}.

    \section{Results and Discussion}
    \noindent
    The simulation results are summarized in Fig.\,\ref{fig:simulation}.
    As expected, the regression scores in Fig.\,\ref{fig:simulation}a,c are similar to the results reported in \cite{sabbagh_predictive_2020}.
    They confirm that the DIAG method is not a consistent estimator for the generative model considered here, 
    and that the RIEMANN method is more robust to pattern noise than SPOC.
    
    The higher robustness to pattern noise of the RIEMANN method generally translated to lower pattern distances compared to SPOC (Fig.\,\ref{fig:simulation}d).
    Regarding the target noise (Fig.\,\ref{fig:simulation}b), SPOC was more robust to higher noise levels, 
    as the distance of the RIEMANN method increased abruptly for $\sigma \ge 1$.
    Note that for $\sigma = 1$ the noise term in (\ref{eq:regmdl}) started to dominate the data term, resulting in poor out-of-sample predictions for both methods (Fig.\,\ref{fig:simulation}a).

\begin{figure*}[t!]
        \centering
        \includegraphics[width=\textwidth]{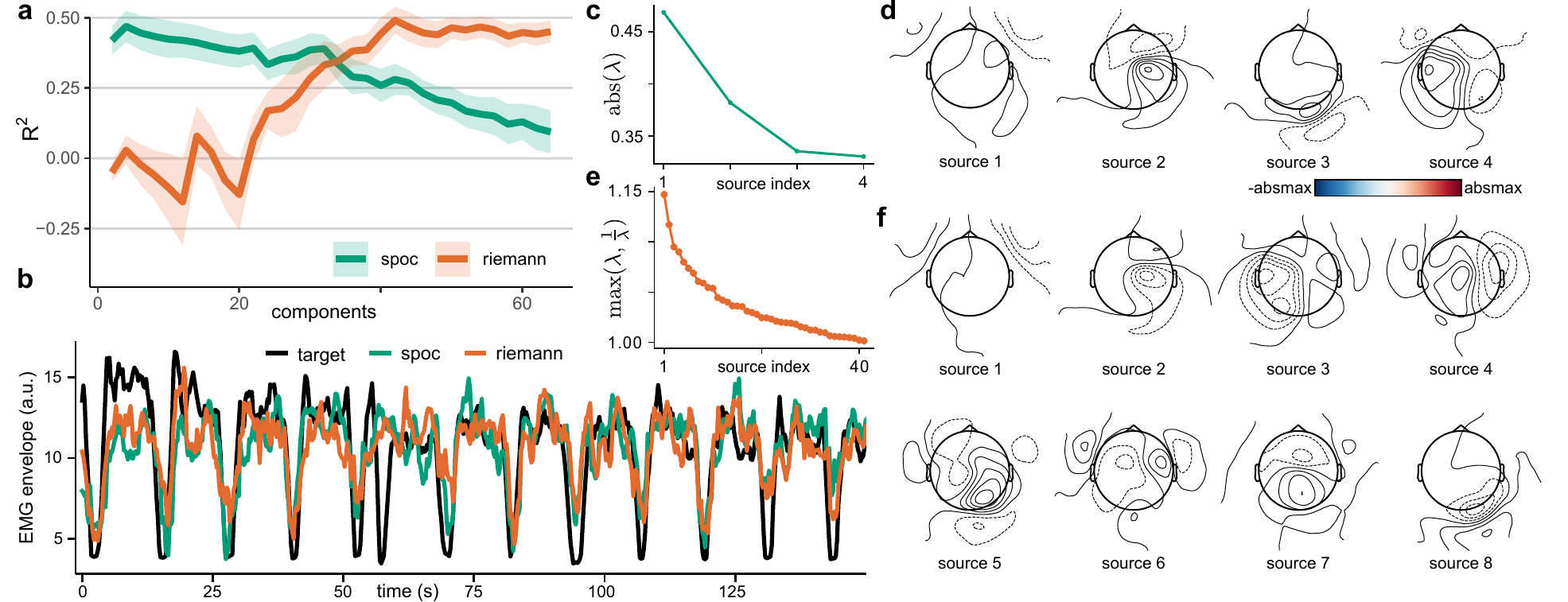}
        \caption{\textbf{Decoding EMG envelope from beta band MEG power.}
                    \textbf{a}, Dependence of the CV test set regression score (explained variance, $R^2$) on the number of components.
                    The number of components was increased from 2 to 64 in steps of 2.
                    The methods are color-coded.
                    Shaded areas show the standard error of the mean.
                    \textbf{b}, Trajectory of the single-channel EMG envelope (black).
                    The test set trajectories of the SPOC and RIEMANN methods are color-coded.
                    \textbf{c}, Eigenvalues ($\lambda$) of the SPOC method with 4 components.
                    The components are sorted in descending order according to the absolute value of the eigenvalue.
                    \textbf{d}, Patterns of the 4 SPOC sources.
                    \textbf{e}, Eigenvalues ($\lambda$) of the RIEMANN method with 42 components.
                    The sources are ordered in descending order according to $\max(\lambda, 1/\lambda)$
                    \textbf{f}, Patterns of the first 8 RIEMANN sources.}
        \label{fig:dscmc}
    \end{figure*}

    Fig.\,\ref{fig:dscmc} summarizes the CMC dataset results for the SPOC and RIEMANN methods.
    Both methods achieved similar quantitative (Fig.\,\ref{fig:dscmc}a) and qualitative (Fig.\,\ref{fig:dscmc}b) decoding accuracies.
    The $R^2$ score peaked at approx. 0.5 (SPOC: 4 components, RIEMANN: 42).
    SPOC reached the peak accuracy at a lower number of components because its components are fitted in a supervised fashion.

    As the patterns in (Fig.\,\ref{fig:dscmc}c-f) indicate, both methods relied on similar sources.
    Considered that the sign is ambiguous, pendants of the 4 SPOC patterns (Fig.\,\ref{fig:dscmc}d) can be readily found among the first 8 RIEMANN patterns (Fig.\,\ref{fig:dscmc}f).
    The first pattern of both methods indicates that they primarily decoded the target from eye artifacts.
    Knowing that the paradigm had a trial-based structure and there was a strong target signal change in the breaks (Fig.\,\ref{fig:dscmc}b), eye artifacts were likely a confounding source.
    This result underlines the importance of interpretable models in M/EEG experiments.	

\begin{figure}[h]
        \centering
        \includegraphics[width=\columnwidth]{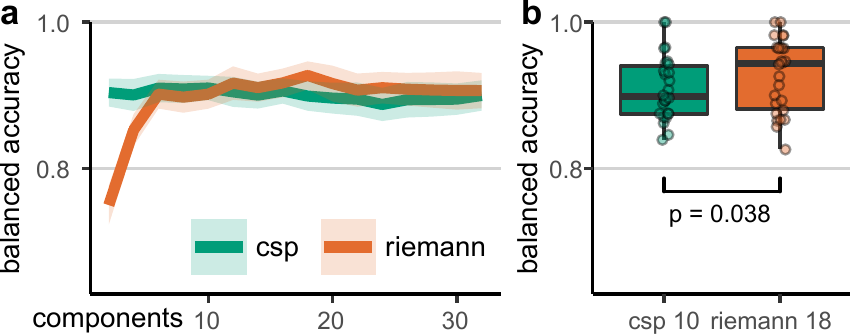}
        \caption{\textbf{Classifying two mental tasks across sessions.}
                    \textbf{a}, Dependence of the classification score (balanced accuracy) on the number of components for a leave one session out CV scheme.
                    The methods are color-coded.
                    Shaded areas show the 95\% confidence interval of the mean.
                    \textbf{b}, Boxplots summarizing the results for the models at the peak in \textbf{a}.
                    The differences between the methods were significant (two-sided, paired t-test, df = 25, $\alpha = 0.05$).}
        \label{fig:dsmims classification}
    \end{figure}

    The multi-session binary classification dataset results are depicted in Fig.\,\ref{fig:dsmims classification}.
    Using 25 sessions to fit the parameters, both methods achieved high accuracies in the test session.
    The peak accuracies for the RIEMANN and CSP methods were 0.93 (18 components per frequency band) and 0.91 (10).
    The paired difference across sessions was significant (Fig.\,\ref{fig:dsmims classification}b).
    
\begin{figure*}[t!]
        \centering
        \includegraphics[width=\textwidth]{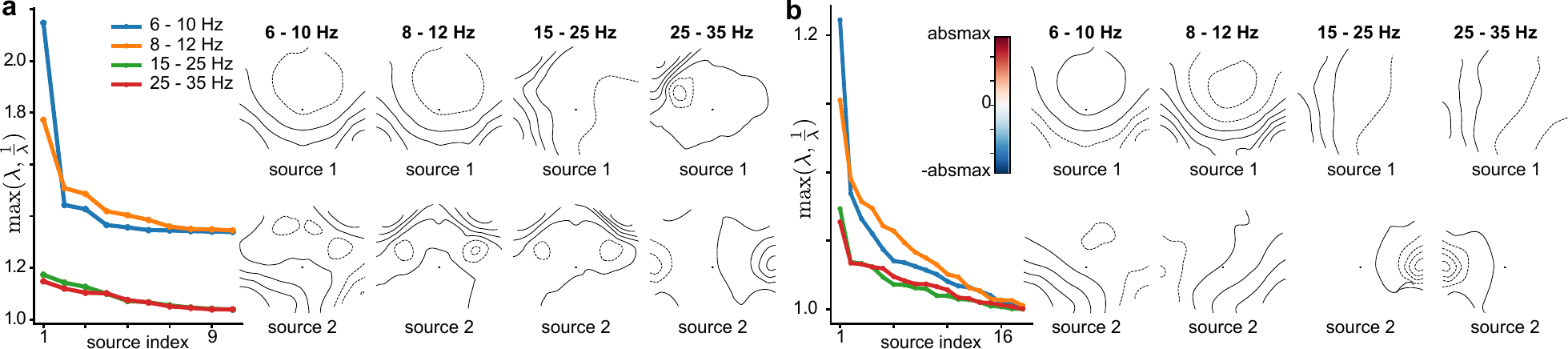}
        \caption{\textbf{Classifying two mental tasks across sessions. Patterns.}
                    \textbf{a}, Eigenvalues (left) and patterns (right) of the CSP method with 10 components per frequency band.
                    The 4 frequency bands in the alpha and beta range are color-coded.
                    For each frequency band, the top 2 patterns are plotted.
                    \textbf{b}, As in \textbf{a} for the RIEMANN method with 18 components.}
        \label{fig:dsmims patterns}
    \end{figure*}

    Fig.\,\ref{fig:dsmims patterns} shows the associated patterns.
    The sources with highest eigenvalues (source 1 in the alpha bands) were similar, 
    indicating that both methods agreed in the most discriminative source.
    The patterns also match with the class-specific, grand-average power modulations, reported in \cite{hehenberger_long-term_2021}.
    We observed two differences.
    First, the RIEMANN patterns were spatially smoother and easier to attribute to single dipolar sources.
    Generally, a higher fraction of dipolar patterns indicates a better source de-mixing quality \cite{delorme_independent_2012}.
    Second, evaluating the eigenvalues in the lower and higher beta band, there was a drastic drop between the first and second eigenvalue for the RIEMANN method.
    This suggests that the first source contained considerably more discriminative information than the second one.
    CSP lacked such a drop, suggesting that it did not identify this beta band source.
    In this longitudinal dataset, the assumption of stationary patterns is certainly violated as the manually mounted electrode cap location varied across sessions.
    Because the electrode locations varied across sessions, both differences could be attributed to the fact that the RIEMANN method is more robust to pattern noise (Fig.\,\ref{fig:simulation}c,d).
    \section{Conclusion}
    \noindent
    We proposed a method to interpret the model parameters of linear regression and classification methods, operating in Riemannian tangent space.
    In simulations, we found that the estimated patterns were robust to noise in the patterns across observations.
    These findings were confirmed in a multi-session EEG dataset.
    The Riemannian tangent space method not only significantly improved the classification accuracy upon CSP but also extracted sources whose patterns were smoother and more dipolar, which are typical characteristics of sources originating in the brain.
    In summary, the proposed approach to compute patterns enables an intuitive interpretation of state-of-the-art linear Riemannian tangent space models.
    \vspace{-6pt}
    \bibliographystyle{IEEEtran}
    \bibliography{references}
    \vspace{-6pt}
    \appendix
    \renewcommand{\theequation}{A.\arabic{equation}}
    \setcounter{equation}{0}
    \begin{small}
    \textit{Proof} that the encoding source patterns $\vec{A}_s$ and unknown regression coefficients $\vec{b}$ can be recovered from the tangent space weight vector $\vec{b}_c$.

    We start the proof with expressing the regression model in (\ref{eq:regmdl}) in terms of the tangent space features at the geometric mean source covariance matrix $\bar{\vec{E}}$.
    Since all $\vec{E}_i$ matrices are diagonal, we have that their geometric mean $\bar{\vec{E}} = \mathrm{diag}(\bar{\vec{p}}) = \mathrm{diag}((\prod_{i=1}^{N} p_{ij})^\frac{1}{N} )$.
    Projecting $\vec{E}_i$ to the tangent space at $\bar{\vec{E}}$ yields:
    \begin{align}
        \tilde{\vec{v}}_i & = \mathrm{proj}_{\bar{\vec{E}}} ( \vec{E}_i ) = \mathrm{upper}( \log( \bar{\vec{E}}^{-\frac{1}{2}} \vec{E}_i \bar{\vec{E}}^{-\frac{1}{2}} ) ) \nonumber\\
        & = \mathrm{upper}( \mathrm{diag}(( \log( p_{ij}) - \log( \bar{p}_j) )_{j=1,..,P} ) )
    \end{align}
    Starting from (\ref{eq:regmdl}), assuming w.l.o.g. that $y$ is zero-mean, and setting $f(p_{ij}) = \log (p_{ij})$ and $b_0 = -\sum_{j=1}^{Q} b_j\,\log(\bar{p}_j)$, it follows that
    \begin{equation}
        y_i = \vec{b}^T f(\vec{p}_i) + b_0 + \varepsilon_i = \underbrace{[ \vec{b}^T, 0 ]}_{\tilde{\vec{b}}^T} \tilde{\vec{v}}_i + \varepsilon_i
        \label{eq:proof:yofv_tilde}
    \end{equation}
    Next, we relate the observed tangent space features $\vec{v}_i$ at $\bar{\vec{C}}$ with $\tilde{\vec{v}}_i$ at $\bar{\vec{E}}$.
    Due to the invariance property of the geometric mean we have $\bar{\vec{C}} = \vec{A} \bar{\vec{E}} \vec{A}^T$. 
    We additionally introduce the matrix $\vec{U} = \bar{\vec{C}}^{\frac{1}{2}} \vec{A}^{-T} \bar{\vec{E}}^{-\frac{1}{2}} $ so that 
    \begin{equation}
       \vec{U}^T \bar{\vec{C}}^{-\frac{1}{2}} \vec{C}_i \bar{\vec{C}}^{-\frac{1}{2}} \vec{U} = \bar{\vec{E}}^{-\frac{1}{2}} \vec{E}_i \bar{\vec{E}}^{-\frac{1}{2}}
    \end{equation}
    holds for all observations $i$.
    It is straightforward to show that $\vec{U}$ is orthogonal. Consequently, we have:
    \begin{align}
        \vec{U}^T \log ( \bar{\vec{C}}^{-\frac{1}{2}} \vec{C}_i \bar{\vec{C}}^{-\frac{1}{2}} ) \vec{U} &= \log(\bar{\vec{E}}^{-\frac{1}{2}} \vec{E}_i \bar{\vec{E}}^{-\frac{1}{2}}) \nonumber\\
        \vec{U}^T \mathrm{upper}^{-1}(\vec{v}_i) \vec{U} &= \mathrm{upper}^{-1}(\tilde{\vec{v}}_i)
    \end{align}
    Now, we can rewrite the dot product between $\tilde{\vec{b}}$ and $\tilde{\vec{v}}$ in (\ref{eq:proof:yofv_tilde}) as:
    \begin{align}
        \tilde{\vec{b}}^T \tilde{\vec{v}}_i &= \mathrm{tr} ( \mathrm{upper}^{-1}(\tilde{\vec{b}}) \mathrm{upper}^{-1}(\tilde{\vec{v}}_i) ) \nonumber\\
        &= \mathrm{tr} ( \mathrm{upper}^{-1}(\tilde{\vec{b}}) \vec{U}^{T} \mathrm{upper}^{-1}(\vec{v}_i) \vec{U} ) \nonumber\\
        &= \mathrm{upper}(\vec{U} \mathrm{upper}^{-1}(\tilde{\vec{b}}) \vec{U}^{T})^T \vec{v}_i = \vec{b}_c^T \vec{v}_i
    \end{align}
    where we defined $\vec{b}_c $ as $ \mathrm{upper}(\vec{U} \mathrm{upper}^{-1}(\tilde{\vec{b}}) \vec{U}^{T})$ and $\mathrm{tr}(\cdot)$ computes the trace of a matrix.
    The weights $\vec{b}_c$ linearly relate the tangent space features $\vec{v}_i$ at $\bar{\vec{C}}$ with the target signal $y_i$.
    Consequently, with $N\rightarrow \infty$ the estimates of a linear estimation method converge to the true weights $\vec{b}_c$.
    
    Since the in-product between the pattern and the weight vector in both tangent spaces is 1 \cite{Haufe2014}, it follows that
    $\vec{d}_c = \mathrm{upper}( \vec{U} \mathrm{upper}^{-1}(\tilde{\vec{d}}) \vec{U}^{T})$
    where $\tilde{\vec{d}} = \tilde{\vec{b}} / ||\tilde{\vec{b}}||^2$ is the pattern associated to $\tilde{\vec{b}}$.
    If we project $\vec{d}_c$ from the tangent space to the covariance matrix space, we get:
    \begin{align}
        \vec{C}_d &= \mathrm{proj}_{\bar{\vec{C}}}^{-1} ( \vec{d}_c ) =  \bar{\vec{C}}^{\frac{1}{2}} \exp ( \mathrm{upper}^{-1}(\vec{d}_c) ) \bar{\vec{C}}^{\frac{1}{2}} \nonumber\\
        &= \bar{\vec{C}}^{\frac{1}{2}} \vec{U} \exp ( \mathrm{upper}^{-1}(\tilde{\vec{d}}) ) \vec{U}^T \bar{\vec{C}}^{\frac{1}{2}} \nonumber\\
        &= \bar{\vec{C}} \vec{A}^{-T} \bar{\vec{E}}^{-\frac{1}{2}}  \exp ( \mathrm{upper}^{-1}(\tilde{\vec{d}}) ) \bar{\vec{E}}^{-\frac{1}{2}} \vec{A}^{-1} \bar{\vec{C}} \nonumber\\
        &= \vec{A}\bar{\vec{E}}^{\frac{1}{2}}  \exp ( \mathrm{upper}^{-1}(\tilde{\vec{d}}) ) \bar{\vec{E}}^{\frac{1}{2}} \vec{A}^T \nonumber\\
        & = \vec{A} \mathrm{diag} \left(\left[ (e^{d_j} \bar{p}_j)_{j=1,..,Q}, (\bar{p}_j)_{j=Q+1,..,P}  \right]\right)  \vec{A}^{T}
    \end{align}
    where we used in the last line the fact that $\mathrm{upper}^{-1}(\tilde{\vec{d}})$ is a diagonal matrix. 
    Computing $\vec{C}_d \bar{\vec{C}}^{-1} $, we get:
    \begin{equation}
        \vec{A} \mathrm{diag} \left(\left[ (e^{\frac{b_j}{||\vec{b}||^2}})_{j=1,..,Q}, (1)_{j=Q+1,..,P}  \right]\right)  \vec{A}^{-1}
    \end{equation}
    Hence, via eigen decomposition of $ \vec{C}_d \bar{\vec{C}}^{-1}$ we can recover the unknown mixing matrix $\vec{A}$ and latent weights $\vec{b}$.
    This concludes the proof.
    
    \end{small}

\end{document}